\documentclass{article}
\usepackage{arxiv}
\usepackage[T1]{fontenc}
\usepackage{microtype}
\usepackage{amsmath,amsfonts}
\usepackage{algorithmic}
\usepackage{algorithm}
\usepackage{array}
\usepackage{tabularx}
\usepackage{subfig}
\usepackage{textcomp}
\usepackage{xurl}
\usepackage{verbatim}
\usepackage{graphicx}
\usepackage[numbers,square]{natbib}
\usepackage[colorlinks=true, linkcolor=blue, anchorcolor=blue, citecolor=blue, urlcolor=blue]{hyperref}
\usepackage{placeins}
\usepackage{makecell}
\usepackage{multirow}
\usepackage{booktabs}
\usepackage{orcidlink}
\usepackage{tikz}
\usepackage{threeparttable}
\newcolumntype{L}[1]{>{\raggedright\arraybackslash}p{#1}}
\newcolumntype{C}[1]{>{\centering\arraybackslash}p{#1}}

\hypersetup{
  pdftitle={Tele-Traversability: Rethinking Traversability for Teleoperated Ground Robots in Terrain Navigation},
  pdfauthor={Lewei Feng, Qi Chen, Wenshuo Wang, Xianghao Meng, Minh Kieu, Haijie Guan, Jin Wu, Junqiang Xi, Fuchun Sun},
  pdfkeywords={Traversability, robotic teleoperation, human-robot interaction, terrain navigation}
}
\begin{document}
\raggedbottom

\title{
  Tele-Traversability: Rethinking Traversability for Teleoperated Ground Robots in Terrain Navigation
}

\author{\parbox{0.94\textwidth}{\centering\bfseries
Lewei~Feng\,\orcidlink{0009-0008-9152-3857}\textsuperscript{1},
Qi~Chen\,\orcidlink{0009-0000-7888-254X}\textsuperscript{1},
Wenshuo~Wang\,\orcidlink{0000-0002-1860-8351}\textsuperscript{2},\\[3pt]
Xianghao~Meng\,\orcidlink{0009-0005-6999-6920}\textsuperscript{1},
Minh~Kieu\,\orcidlink{0000-0001-7798-6195}\textsuperscript{3},
Haijie~Guan\,\orcidlink{0000-0001-9046-6944}\textsuperscript{1},\\[3pt]
Jin~Wu\,\orcidlink{0000-0001-5930-4170}\textsuperscript{4},
Fuchun~Sun\,\orcidlink{0000-0003-3546-6305}\textsuperscript{5},
and Junqiang~Xi\,\orcidlink{0000-0001-8607-4542}\textsuperscript{2}\\[7pt]
\normalfont\small
\textsuperscript{1}School of Mechanical Engineering, Beijing Institute of Technology, Beijing 100081, China\\
\textsuperscript{2}Faculty of Marine Science and Technology, Beijing Institute of Technology, Zhuhai 519088, China\\
\textsuperscript{3}Department of Civil and Environmental Engineering, University of Auckland, Auckland 1010, New Zealand\\
\textsuperscript{4}School of Artificial Intelligence, University of Science and Technology Beijing, China\\
\textsuperscript{5}Department of Computer Science, Tsinghua University, China\\[5pt]
Corresponding authors: Wenshuo Wang; Junqiang Xi\\
\href{mailto:ws.wang@bit.edu.cn}{\nolinkurl{ws.wang@bit.edu.cn}}; \href{mailto:xijunqiang@bit.edu.cn}{\nolinkurl{xijunqiang@bit.edu.cn}}
\thanks{Other author contacts: Lewei Feng, \nolinkurl{3120245245@bit.edu.cn}; Qi Chen, \nolinkurl{2723821719@qq.com}; Xianghao Meng, \nolinkurl{3120235085@bit.edu.cn}; Minh Kieu, \nolinkurl{minh.kieu@auckland.ac.nz}; Jin Wu, \nolinkurl{wujin@ustb.edu.cn}; Fuchun Sun, \nolinkurl{fcsun@tsinghua.edu.cn}.}
}}
\date{}
\renewcommand{\shorttitle}{Tele-Traversability}

\fancyhead{}
\renewcommand{\headrulewidth}{0pt}

\maketitle

\begin{abstract}
Teleoperation allows a human operator to remotely command and guide a mobile robot to navigate in off-road environments, yet fluent and user-friendly tele-navigation requires an alignment of traversability evaluation between the operator and the robot. The human operator typically utilizes \textit{off-site} incomplete and delayed feedback via a tailored user interface to make a judgment of traversability, while the robot makes such an evaluation based on \textit{in situ} onboard sensory information, which could cause divergent traversability estimation, generating misaligned decisions and actions. Current methods for traversability modeling, estimation, and prediction are mainly derived from the view of robots (i.e., robot-centric), suitable for fully autonomous mobile robots, but neglect the influence of human operators. To bridge this gap, this survey for the first time extends the concept of traversability from robot-centric to human-centric by accounting for the operator's cognition states (e.g., attention, workload, and risk tolerance/awareness), termed as \textit{tele-traversability}. We first revisit the traversability definitions and roles in robotics and then extend them to teleoperation contexts. Finally, we highlight the future trends and open challenges of tele-traversability toward human-centric teleoperation systems.

\end{abstract}


\section{Introduction}
\begin{figure}[t]
  \centering1
  \includegraphics[scale=1.0]{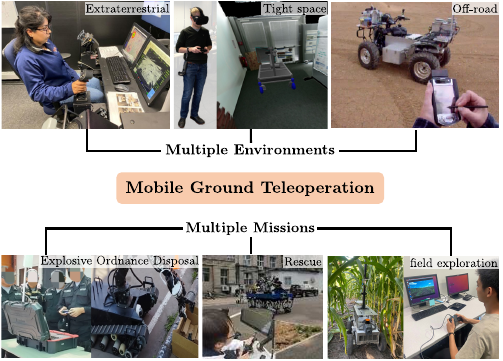}
  \caption{Applications of mobile ground teleoperation\cite{litakerLunarTerrainVehicle2025, stotkoVRSystemImmersive2019,  fongPdaDriverHandheldSystem, canazaccariJVC02TeleoperatedRobot2024, zhengTeleoperatedMobileRobotic2024, chakrabortyRealTimeGenerationDelayCompensated2025}.}
  \label{fig_0}
\end{figure}

Teleoperation of ground robots, a human controlling a robot at a distance, is critical for navigating in unknown and complex terrains where fully autonomous is impractical or undesirable (Fig. \ref{fig_0}). The human operator must acquire sufficient awareness of the terrains via a limited interface (e.g., displays, haptics) and then leverage their prior domain knowledge and observations to make decisions about \textit{whether} the robot can safely go through the region (i.e., decision-making), \textit{where} the robot should navigate (i.e., planning), and \textit{what} action the robot should take (i.e., action). A user-friendly teleoperated ground robot should be able to understand the operator's intent and then make a desirable decision and plan a trackable path for terrain navigation. With advances in artificial intelligence and sensory techniques, robots could execute tasks (e.g., terrain perception, short-range path planning) along this pipeline while collaborating with humans. When navigating in terrains, the robot should first \textit{assess} whether it can travel the approaching terrain without causing damage or getting stuck. 

\begin{figure}[h]
  \centering
  \includegraphics[width=\linewidth,height=0.60\textheight,keepaspectratio]{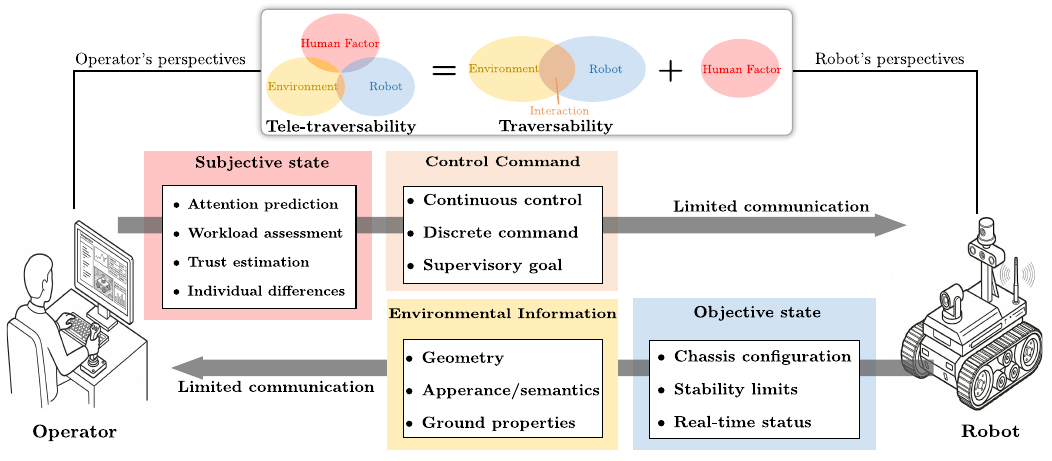}
  \caption{\textbf{Tele-traversability} extends classical traversability by explicitly incorporating the operator.}
  \label{fig_3}
\end{figure}

\begin{figure}[t]
  \centering
  \includegraphics[scale=1.0]{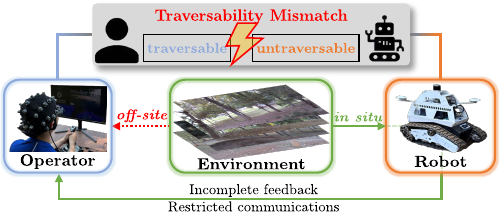}
  \caption{The traversability mismatch between robot and operator.}
  \label{fig_1}
\end{figure}

\textbf{Traversability} refers to the ability of a ground robot to enter and cross a given terrain region under constraints, as quantified by terrain-robot interplay and admissible-state optimality \cite{papadakisTerrainTraversabilityAnalysis2013}. Traversability estimation is a prerequisite to provide risk evaluation during terrain navigation \cite{papadakisTerrainTraversabilityAnalysis2013,sevastopoulosSurveyTraversabilityEstimation2022, borgesSurveyTerrainTraversability2022, beycimenComprehensiveSurveyUnmanned2023, shuOverviewTerrainTraversability2025} for ground vehicles, planetary rovers, and other mobile platforms \cite{serajiTraversabilityIndexNew1999, fankhauser2014robot, fanLearningRiskAwareCostmaps2022, dixitSTEPStochasticTraversability2024}. For fully-autonomous ground robots, traversability is typically used to bridge perception and decision by interpreting terrain feasibility and risk through geometric (e.g., slope), appearance (e.g., color), or semantic cues (e.g., water, grass) from environments into actionable representations \cite{shuOverviewTerrainTraversability2025, fanLearningRiskAwareCostmaps2022}, providing a self-contained basis of motion planning \cite{borgesSurveyTerrainTraversability2022, mengTerrainNetVisualModeling2023}. While this works for autonomous robots, it may fail for ground robots \textit{teleoperated} by a human operator. Therefore, an ideal estimated traversability for teleoperated systems should properly integrate the operator's expectation (Fig.~\ref{fig_3}) with the robot's \textit{in situ} estimation. For instance, the robot should make a safe decision and plan a traversable route that is as close as possible to its operator's intent; otherwise, human-robot conflicts may occur (Fig.~\ref{fig_1}). 

Unlike traversability in fully-autonomous robots, here we introduce \textbf{tele-traversability}, a human-centric metric to systematically assess whether a teleoperated ground robot \textit{and} its human operator \textit{jointly} deem a region traversable. This reframes the core question from ``Can the robot traverse this region?'' to ``Do the operator and robot agree on traversability?'' Tele-traversability depends on triadic alignment among the environment, robot, and operator (Fig.~\ref{fig_3}). The robot perceives the environment via onboard sensors, generates paths compatible with the operator's cognitive model while delivering critical, user-expected information. Thus, traversability is no longer a purely robot-centric metric for onboard planning; instead, it becomes a hybrid performance measure that balances the robot's physical feasibility with the operator's perceptual acceptability. This perspective enables human-centric teleoperation system design, moving beyond the traditional approaches that treat communication uncertainty, conflict detection, and authority allocation in isolation. Tele-traversability supports operator guidance, intervention triggering, and shared decision-making across autonomy regimes, enabling seamless human-robot coordination \cite{fong2001safeguarded, draganPolicyblendingFormalismShared2013, goodrichTeleoperationAssistiveHumanoid2013}.

\begin{figure}[t]
  \centering
  \includegraphics[width=\linewidth,height=0.60\textheight,keepaspectratio]{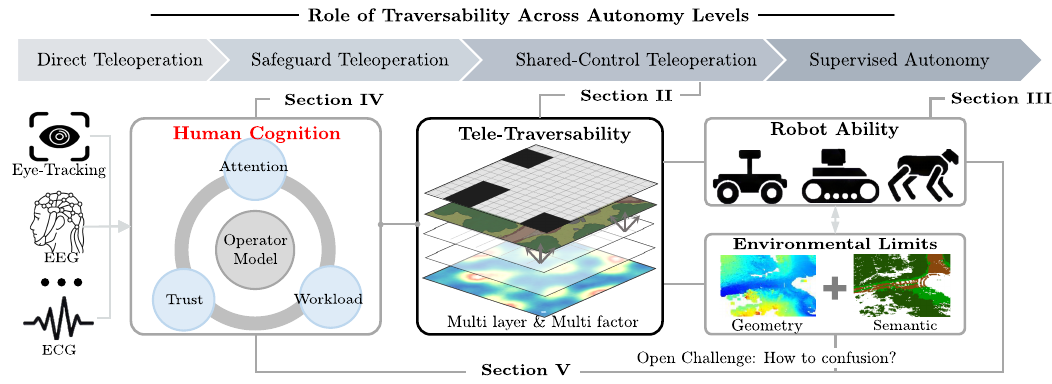}
  \caption{Paper organization and conceptual map. (Some icons designed by Gemini 3 Pro.) }
  \label{fig_2}
\end{figure}

Significant progress has been made in either traversability estimation \textit{or} robot teleoperation, yet their intersection remains relatively underexplored. Researchers inferred traversability from geometric \cite{helmick2009terrain, molinoTraversabilityMetricsRough2007, heckmanPotentialNegativeObstacle2007}, interaction, appearance, and semantic cues \cite{ishigami2006terramechanics, reinaSlipbasedTerrainEstimation2016, bajracharyaLearningLongrangeTerrain2008, mcdanielTerrainClassificationIdentification2012}, create a reasoning paradigm with rule/physics-based and \cite{ishigami2006terramechanics, molinoTraversabilityMetricsRough2007} data-driven models \cite{martinezSupervisedLearningNaturalTerrain2020, zurnSelfSupervisedVisualTerrain2021, freyRoadRunnerLearningTraversability2024}, and learn actionable representations for downstream modules (e.g., decision-making and path planning) \cite{dahlkampSelfsupervisedMonocularRoad2006, ahtiainenAugmentingTraversabilityMaps2013, fankhauserProbabilisticTerrainMapping2018, sivaSelfReflectiveTerrainAwareRobot2021, caiEVORADeepEvidential2024}. However, most of them treat traversability as a \textit{robot-centric} metric based on onboard sensing \cite{borgesSurveyTerrainTraversability2022,sevastopoulosSurveyTraversabilityEstimation2022, tejiSurveyOffRoadMobile2023, beycimenComprehensiveSurveyUnmanned2023, shuOverviewTerrainTraversability2025}, linking terrain features directly to the robot's mobility performance. Literature on teleoperation primarily focuses on communication constraints, human-robot interfaces, and shared control \cite{javdaniSharedAutonomyHindsight2018, selvaggioSharedControlTeleoperationArchitecture2022, gopinathHumanintheLoopOptimizationShared2017}, but neglects the underlying mobility representation. To bridge this gap in \cite{livatinoIntuitiveRobotTeleoperation2021, shillehBestWorstExternal2021, pencoAnticipatoryAdaptiveFootstep2025}, we extend it to teleoperated robots in terrain navigation by introducing the notion of \textit{tele-traversability} (Fig.~\ref{fig_2}). Our scope excludes general motion planning, low-level control, and terrain classification lacking explicit traversability analysis. We provide a taxonomy of tele-traversability (Section~2), revisit traversability estimation methods and representations (Section~3), and take a tele-traversability analysis (Section~4). We conclude by outlining open challenges toward a unified framework for tele-robotics terrain navigability.

\section{From Traversability to Tele-Traverability: Definitions and Roles}
Most prior work defines traversability from a robot-centric perspective by assessing whether a terrain supports mobile robot motion \cite{shuOverviewTerrainTraversability2025}. However, interpretations vary based on how traversal feasibility is quantified (e.g., passability, cost, probability, or risk) and on the requirements of downstream planning and control \cite{ederTraversabilityAnalysisOffroad2023}. This formulation becomes insufficient when a human operator is involved, as traversability depends not only on autonomous feasibility but also on the operator's perceived acceptability of traversal under the situational context. To address this ambiguity, we introduce a unified taxonomy by formalizing robot-centric traversability and extending it to \emph{tele-traversability},  a concept integrating human judgment into traversability assessment. While related terms such as \textit{drivability}, \textit{trafficability}, \textit{navigability}, \textit{mobility}, and \textit{maneuverability} are widely used \cite{papadakisTerrainTraversabilityAnalysis2013}, we adopt \textit{traversability} as an umbrella term in this survey for consistency.

\subsection{Traversability in Fully Autonomous Navigation}\label{section:2.1}

Traversability is not an inherent property of terrain alone but rather a relational assessment that depends on the interaction between the terrain and the robot's mobility capabilities. This assessment involves three key components: 
\begin{itemize}
    \item A \textbf{terrain model} of characterizing environmental properties (e.g., geometry, texture, or deformability),
    \item A \textbf{robot model} that defines the platform's mobility constraints and motion capabilities, and 
    \item \textbf{Evaluation criteria} of quantifying feasibility, safety margin, or risk based on terrain-robot interaction.
\end{itemize}

\subsubsection{Terrain Model}
Terrain models are essential for assessing traversability by capturing the environmental factors such as geometric, physical, and semantic attributes of the support surface \cite{goodinFastTerrainTraversability2021, vandapelNaturalTerrainClassification2004}. Geometric information alone is often inadequate for predicting robot-terrain interactions, modern representations integrate these heterogeneous features for more robust estimation \cite{ xue2023traversability, shuOverviewTerrainTraversability2025}. Recent advances further incorporate uncertainty and temporal dynamics, enabling probabilistic models that support risk-aware decision-making in partially observable dynamic environments \cite{fankhauserProbabilisticTerrainMapping2018, dixitSTEPStochasticTraversability2024, knaupSafeHighPerformanceAutonomous2023}.

\subsubsection{Robot Model}
The robot model is fundamental to traversability assessment, as it encodes mobility constraints, motion capabilities, an terrain interaction dynamics \cite{beycimenComprehensiveSurveyUnmanned2023, borgesSurveyTerrainTraversability2022}. By representing the robot's physical limits, the model determines feasible traversable regions, stability conditions, and operational sustainability. These constraints are usually formalized through kinematic, dynamic, or kinodynamic models, incorporating factors such as motion geometry, traction, slip, and stability. Such models enable the estimation of key traversability metrics like negotiable slope angles, rollover margins, and mobility failure conditions. In adaptive systems, real-time parameter updates ensure predictions remain accurate under varying terrain conditions \cite{reinaSlipbasedTerrainEstimation2016, sivaSelfReflectiveTerrainAwareRobot2021}.

\subsubsection{Evaluation Criteria}
Traversability evaluation criteria determine how terrain conditions and robot capabilities are assessed for safe and effective navigation \cite{sevastopoulosSurveyTraversabilityEstimation2022, borgesSurveyTerrainTraversability2022}. These criteria classify terrain as \textit{acceptable}, \textit{risky}, or \textit{infeasible} for traversal. While some assessments rely solely on terrain properties (e.g., extreme slopes or roughness), most integrate environmental demands with the robot's physical limits. In practice, traversability is often quantified through unified metrics, such as scores, cost functions, or decision variables, that balance safety, efficiency, and motion feasibility \cite{dixitSTEPStochasticTraversability2024, knaupSafeHighPerformanceAutonomous2023}. The choice of criteria reflects different interpretative frameworks. Some approaches enforce strict boundaries, rejecting terrain if estimated demands exceed robot capabilities \cite{belluttaTerrainPerceptionDEMO2000, hamnerEfficientSystemCombined2008}. Others employ continuous costs or mobility indices to represent traversal difficulty rather than binary feasibility \cite{goodinFastTerrainTraversability2021, wuRealtimeLiDARbasedTerrain2025}. More advanced approaches extend beyond feasibility to assess low-probability, high-consequence risks (e.g., slip or rollover) using probabilistic or risk-aware formulations \cite{fankhauserProbabilisticTerrainMapping2018, fanLearningRiskAwareCostmaps2022, benrabahReviewTraversabilityRisk2024}. 

\subsection{Tele-traversability in Teleoperated Robots}
\label{section:2.2}
In teleoperation, robot-centric traversability becomes insufficient under partial autonomy. Safe traversal no longer depends solely on the robot's capabilities, as perception, decision-making, and control are shared between the robot and a remote operator \cite{fong2001safeguarded, darvishTeleoperationHumanoidRobots2023, kohnMeasurementTrustAutomation2021}. This shared autonomy redefines traversability into \textit{tele-traversability}, a joint assessment shaped by terrain properties, the robot's mobility constraints, and the operator's perception and risk tolerance, all mediated by limited sensory feedback. Crucially, traversability is no longer just a robot's internal passability estimate but a negotiated outcome of human-robot collaboration \cite{javdaniSharedAutonomyHindsight2018, verhagenMeaningfulHumanControl2024}. Thus, tele-traversability comprises two key aspects: 
\begin{itemize}
    \item \textbf{Robotic feasibility}: Whether the robot can physically and dynamically traverse the region while adhering to kinematic/dynamic constraints, environmental uncertainty, and predefined risk thresholds.
    \item \textbf{Operator acceptability}: Whether the traversal aligns with the operator's intent, perceptual/cognitive limits, and risk tolerance, given teleoperation interfaces and situational awareness (Fig. \ref{fig_3}). 
\end{itemize}
Tele-traversability is dynamically shaped through human-robot interaction, with its final assessment, for instance, their disparities of risk awareness, attention allocation, and risk tolerance. Based on the autonomy level and application, tele-traversability can be quantified as scalar scores or structured metrics. For functional relevance, these metrics should convey estimated confidence, highlight potential human-robot disagreement, and determine whether explanatory feedback, intervention, or authority reallocation is necessary. However, current modeling approaches fail to fully capture the tele-traversability. For instance, physics-based terrain-robot interaction models quantify embodied mobility constraints \cite{borgesSurveyTerrainTraversability2022}, probabilistic methods formalize uncertainty and risk \cite{dixitSTEPStochasticTraversability2024}, learning-based perception improves terrain interpretation, and human behavioral models characterize operator states and responses \cite{carissoliMentalWorkloadHumanRobot2024}.  

Recent advances in \textit{foundation models} present a promising solution to bridge this gap. These large language and multimodal models exhibit robust cross-modal reasoning, context aggregation, and task-transfer knowledge \cite{brownLanguageModelsAre2020, driessPaLMEEmbodiedMultimodal2023, openaiGPT4TechnicalReport2024}. Early applications in teleoperation and embodied robotics leverage them for intent inference, adaptive strategy generation, and human-robot-environment coordination \cite{feiLargeLanguageModelAidedAssistiveRobot2025, taoLAMSLLMDrivenAutomatic2025, ahnAutoRTEmbodiedFoundation2024}. The key advantage of tele-traversability lies in unifying three traditionally disjoint judgments of environmental comprehension, robotic feasibility, and operator intent/acceptance.

\subsubsection{Terrain and Robot Models}
Robot interacts directly with terrain, whereas human operators engage with the environment through a (virtual) tailored user-interface. This distinction ensures that conventional traversability analysis frameworks remain valid. These frameworks assess terrain features (e.g., slope gradient, surface roughness) as kinematic constraints, while robot-specific feasibility depends on embodied dynamics, motion limitations, and terrain interaction mechanics  \cite{ederTraversabilityAnalysisOffroad2023, beycimenComprehensiveSurveyUnmanned2023}. 

While conventional physical modeling captures robot-environment interactions, characterizing the robot itself requires a multidimensional approach beyond purely mechanical evaluation. Robotic systems must traverse terrain while performing assistive and communicative functions with human operators. Thus, tele-traversability requires an expanded modeling framework that integrates: (i) autonomous competency, (ii) functional boundaries of autonomy, and (iii) real-time authority allocation between humans and robots \cite{gopinathHumanintheLoopOptimizationShared2017}. This framework must account for both physical constraints (e.g., embodiment, kinematics) and autonomy-related factors, such as onboard traversability assessment, decision responsibility allocation \cite{nielsenEcologicalInterfacesImproving2007, livatinoIntuitiveRobotTeleoperation2021}, and dynamic adaptation to task- and context-dependent autonomy \cite{darvishTeleoperationHumanoidRobots2023}. These factors shape the necessary human-robot coordination dynamics \cite{selvaggioSharedControlTeleoperationArchitecture2022, carissoliMentalWorkloadHumanRobot2024}. Section \ref{section:2.3} provides a detailed analysis of these autonomy-dependent variations.

\begin{table}[!t]
\centering
\caption{Preliminary evaluation dimensions for tele-traversability.}
\label{tab_1}
\begin{threeparttable}
\footnotesize
\setlength{\tabcolsep}{4.0pt}
\renewcommand{\arraystretch}{1.16}
\begin{tabularx}{\textwidth}{L{0.22\textwidth} C{0.075\textwidth} L{0.24\textwidth} >{\raggedright\arraybackslash}X C{0.085\textwidth}}
\toprule
\toprule
\textbf{Dimension} & \textbf{Symbol} & \textbf{Canonical form} & \textbf{Meaning} & \textbf{Typical mode} \\
\midrule

\textbf{Operator-side acceptability}
& $A_{\mathrm{acc}}$
& $\dfrac{1}{N}\sum_{i=1}^{N}\mathbf{1}\!\left(r_i\in\Omega_o\right)$
& Fraction of robot-side traversability or risk estimates $r_i$ that fall within the acceptable region $\Omega_o$ of operator $o$.
& O/P \\

\textbf{Human--robot agreement}
& $A_{\mathrm{agr}}$
& $1-\widetilde{\mathcal D}(\mathbf{h},\mathbf{r})$
& Agreement between human and robot judgment vectors $\mathbf{h}$ and $\mathbf{r}$, where $\widetilde{\mathcal D}(\mathbf{h},\mathbf{r})$ denotes a normalized disagreement measure.
& O/P \\

\textbf{Borderline threshold alignment}
& $A_{\mathrm{bnd}}$
& $\max\!\left(0,\;1-\dfrac{|\tau_h-\tau_r|}{\Delta\tau}\right)$
& Degree to which human $\tau_h$ and robot $\tau_r$ exhibit similar decision thresholds in the borderline or decision-critical regions, where $\Delta\tau$ is the normalization range.
& P \\

\textbf{Representation intelligibility}
& $A_{\mathrm{int}}$
& $\dfrac{1}{K}\sum_{k=1}^{K}\mathbf{1}\!\left(\hat{y}_k = y_k\right)$
& Proportion of traversability representations for which the operator can correctly interpret the robot-side judgment, where $\hat{y}_k$ and $y_k$ denote the operator's interpretation and the reference meaning for frame $k$, respectively.
& P \\

\textbf{Personalization benefit}
& $G_{\mathrm{pers}}$
& $\dfrac{S_{\mathrm{pers}}-S_{\mathrm{gen}}}{|S_{\mathrm{gen}}|+\epsilon}$
& Relative benefit of a personalized model over a generic model, where $S_{\mathrm{pers}}$ and $S_{\mathrm{gen}}$ denote performance under the same downstream score $S$.
& P \\

\textbf{Downstream performance}
& $U_{\mathrm{down}}$
& $\dfrac{1}{M}\sum_{j=1}^{M}u_j$
& Aggregated downstream performance, where $u_j$ denotes the normalized coordination outcome for episode or evaluation unit $j$.
& O/P \\

\bottomrule
\bottomrule
\end{tabularx}

\begin{tablenotes}[flushleft]
\scriptsize
\item \textit{Notation}: 
$N$, $K$, $M$ denote the numbers of evaluated items, intelligibility items, and downstream coordination episodes, respectively; 
$\mathbf{1}(\cdot)$ is the indicator function; 
$\epsilon>0$ is a small constant for numerical stability.

\item \textit{Typical mode}: O = online; P = post hoc; O/P = either online or post hoc, depending on system design and study protocol.
\end{tablenotes}
\end{threeparttable}
\end{table}

\subsubsection{Operator Models}
In teleoperation systems, the operator influences task performance solely through remote interaction with the robotic platform, lacking direct physical contact with the environment. As a result, environmental awareness depends entirely on the robot's feedback \cite{selvaggioSharedControlTeleoperationArchitecture2022}. Effective tele-traversability evaluation must therefore consider two cognitive processes: (i) the operator's perceptual reconstruction of terrain features from robotic sensory data, and (ii) the translation of this perception into control commands. To formalize this linkage, we propose quantifying the operator's cognitive state via measurable human factors. This elucidates the psychomotor mechanisms behind decision-making, enabling their integration into traversability assessment algorithms and adaptive human-robot interaction frameworks \cite{verhagenMeaningfulHumanControl2024}.

Effective teleoperation requires modeling the operator's cognitive processes, such as their perception of salient terrain cues,  parallel information processing capacity, and trust in the robot's reliability. Critical human factors (e.g., attention allocation, workload, and cognitive readiness) directly shape the operator's judgment and control in mediated interactions \cite{nielsenEcologicalInterfacesImproving2007, carissoliMentalWorkloadHumanRobot2024}. Because these mechanisms adapt to task demands, interface conditions, and context, they should be treated as a dynamic component of tele-traversability rather than static operator traits. 

Operator experience and individual differences also influence tele-traversability assessments \cite{wangDrivingStyleClassification2017, livatinoIntuitiveRobotTeleoperation2021}. For instance, experienced operators tend to predict robot-terrain interactions more accurately, keep situational awareness despite limited feedback, and seamlessly integrate onboard autonomy. In contrast, novice operators often rely heavily on explicit feedback and adopt more conservative strategies, especially under uncertainty or perceptual limitations. As a result, identical terrain conditions, robot states, and autonomy support may lead to divergent decisions across operators. Effective operator modeling must therefore account for these persistent individual differences in traversability assessment and control strategies \cite{gopinathHumanintheLoopOptimizationShared2017}. A deeper discussion follows in Section \ref{section:4.2}.

\subsubsection{Evaluation Criteria}
Extending traversability to tele-traversability requires adapting the evaluation criteria defined in Section \ref{section:2.1}. Beyond assessing physical plausibility and risk-awareness in autonomous perception, the framework must prioritize dynamic alignment between robotic assessment and operator acceptability during execution. For instance, operators may reject robot-classified traversable terrain due to perceived uncertainty or overriding robotic warnings based on experience. This shift places human-robot alignment on par with physical risk evaluation \cite{javdaniSharedAutonomyHindsight2018, gopinathHumanintheLoopOptimizationShared2017}. As outlined in Table~\ref{tab_1}, the augmented framework introduces five key dimensions: operator acceptability, human-robot judgment concordance, threshold alignment, representation interpretability, and personalization efficacy. These metrics preserve conventional traversability measures while addressing the need for human-aligned decision support in teleoperation.

In addition, evaluation must extend beyond representational fidelity. In shared autonomy and teleoperation, a system's true value lies not in perceptual accuracy but in its impact on human-robot coordination. Empirical studies show that system efficacy is better measured by task efficiency, collision rates, cognitive load, intervention dynamics, and perceived authority, metrics that often reveal critical trade-offs overlooked by perception assessments \cite{javdaniSharedAutonomyHindsight2018, selvaggioSharedControlTeleoperationArchitecture2022}. Thus, tele-traversability frameworks should be assessed not only on their ability to model terrain, kinematics, and operator constraints but on their capacity to detect human-robot goal misalignment and improve coordination through integration.

\begin{figure}
  \centering
  \includegraphics[scale=1.0]{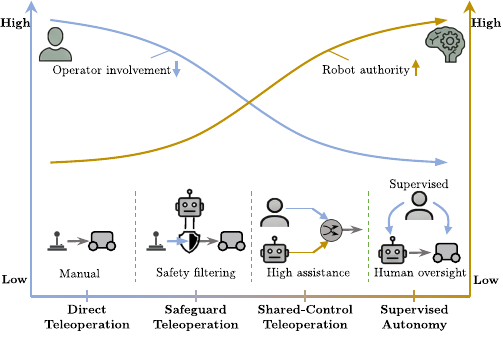}
  \caption{Traversability’s role across autonomy levels in teleoperation.}
  \label{fig_8}
\end{figure}

\subsection{Tele-Traversability at Different Autonomy Levels}
\label{section:2.3}
Modern taxonomies frame autonomy as a dynamic spectrum, where perception, decision-making, and control authority are fluidly distributed between human operators and robots \cite{on2021taxonomy, ireguiReconfigurableConstraintBasedReactive2021}. Build on teleoperation and shared autonomy research \cite{darvishTeleoperationHumanoidRobots2023, fongAdvancedInterfacesVehicle2001, goodrichHumanRobotInteraction2008, fong2001safeguarded, goodrichTeleoperationAssistiveHumanoid2013, fongRobotAskerQuestions2003}, we identify four operational modes (Fig. \ref{fig_8}): \textit{direct}, \textit{safeguarded}, \textit{shared-control} teleoperation, and \textit{supervised autonomy}. These modes represent points along a continuum rather than discrete categories. As shown in \cite{beerFrameworkLevelsRobot2014, methnaniWhosChargeHere2024}, real-world systems often blend these modes, adapting authority based on environmental complexity, robot capabilities, and operator workload. While traversability remains rooted in terrain-vehicle physical interactions  \cite{borgesSurveyTerrainTraversability2022, beycimenRealTimeTerrainTraversability, shuOverviewTerrainTraversability2025}, its implementation must align with the temporal and contextual demands of each autonomy mode.

\emph{Direct teleoperation.} 
In direct teleoperation, operators send low-level control commands (e.g., velocity, steering inputs), while the robot transmits sensor feedback without autonomous intervention. This mode is typical in classical teleoperation systems for ground vehicles, which rely on real-time visual data (e.g., camera feeds, depth maps) or basic occupancy representations but do not autonomously modify trajectories \cite{darvishTeleoperationHumanoidRobots2023, fongAdvancedInterfacesVehicle2001, chilianStereoCameraBased2009}. Here, traversability analysis primarily enhances perception; hazard markers, binary traversability overlays, or terrain classification may be overlaid on visual or geometric representations (e.g., images, point clouds, elevation maps) to improve situational awareness \cite{beycimenRealTimeTerrainTraversability, shuOverviewTerrainTraversability2025, milellaSelflearningFrameworkStatistical2015, de2009survey, reinaTerrainAssessmentPrecision2017}. These visualizations aid hazard identification but impose no algorithmic constraints on control, as the operator remains in full authority  \cite{goodrichTeleoperationAssistiveHumanoid2013, chilianStereoCameraBased2009, milellaSelflearningFrameworkStatistical2015}. Thus, tele-traversability serves as a decision-support tool rather than a shared autonomy feature, maintaining manual control while improving environmental comprehension.

\emph{Safeguarded teleoperation.} 
In this mode, traversability assessment not only informs the operator but also actively enforces safe constraints. While human commands remain the primary input, they are conditionally executed; that is, the robot verifies their safety in real-time via onboard perception and suppresses those that would lead to hazardous actions \cite{fong2001safeguarded, fongRobotAskerQuestions2003}. Typical safeguards include collision avoidance, minimum hazard-distance enforcement, and stability-preserving motion constraints. Operator override authority (i.e., forcibly executing rejected commands) is implementation-dependent and dictated by task requirements. Here, traversability serves a \textit{dual} role: providing operator-facing situational awareness (as in direct teleoperation) while enforcing non-overridable safety boundaries. These boundaries trigger reactive measures such as command rejection, forced stops, or exclusion-zone enforcement. This distinguishes safeguard teleoperation from direct control; that is, traversability is not just visualized but actively imposed by systems. However, the system does not achieve \textit{shared autonomy}: it neither infers operator intent nor generates corrective trajectories when commands are blocked. Interventions remain \textit{local} and \textit{reactive}, limited to predefined safety violations. Thus, safeguarded teleoperation occupies an intermediate niche between direct teleoperation and shared control \cite{fongRobotAskerQuestions2003}.

\emph{Shared-control teleoperation.} 
As robots grows more autonomous, their authority in shared-control systems expands. These systems leverage the robot's capability to contribute to motion generation, blending operator input with autonomous execution. A common paradigm involves the operator issuing high-level commands (e.g., direction or intent) while the robot handles low-level control under physical constraints \cite{goodrichHumanRobotInteraction2008, javdaniSharedAutonomyHindsight2018}. Tele-traversability actively shapes the robot’s decision-making process\cite{draganPolicyblendingFormalismShared2013, javdaniSharedAutonomyHindsight2018}. However, integrating traversability into autonomy may create conflicts, for instance, the robot may override the operator's reasonable commands. Such misalignment between human judgment and robotic constraints can erode trust, trigger overrides, and degrade their collaboration \cite{draganPolicyblendingFormalismShared2013, javdaniSharedAutonomyHindsight2018, gopinathHumanintheLoopOptimizationShared2017}. Tele-traversability mitigates these by mediating between robot-centric assessment (e.g., terrain feasibility, risk models) and operator-centric considerations (e.g., terrain perception, risk tolerance). It serves a \textit{dual} role: 
\begin{itemize}
    \item \textbf{Assistive autonomy.} Instead of merely evaluating safety, tele-traversability aligns robotic assistance with operator expectations, ensuring that interventions are physically viable and cognitively compatible. This prevents disruptive deviations from the operator's mental model, which is critical for shared control to feel intuitive rather than obstructive.
    \item \textbf{Authority allocation.} Tele-traversability provides a principled metric to dynamically adjust human-robot control. Existing methods employ fixed role divisions or adaptive adjustments based on risk, uncertainty, or workload \cite{draganPolicyblendingFormalismShared2013, gopinathHumanintheLoopOptimizationShared2017}. By evaluating terrain context, the system determines whether the current authority balance remains suitable.  If not, it modulates the operator's input by preserving, adapting, or supplementing it as needed.
\end{itemize}


\emph{Supervised autonomy.} 
Supervised autonomy represents the highest level of teleoperation, where the robotic system handles primary navigation while the human operator assumes a supervisory role, such as configuring policies and intervening in exceptional cases. This aligns with established automation frameworks that recast human operators from direct control to high-level decision-making and anomaly resolution \cite{parasuramanModelTypesLevels2000, beerFrameworkLevelsRobot2014, on2021taxonomy}. In practice, interaction shifts from continuous control to intermittent supervision, where operators specify mission objectives while the autonomous system executes tasks, requesting human input only when necessary. Within this framework, tele-traversability evolves from enabling direct control to being embedded in the autonomy framework. For off-road navigation, traversability analysis is integrated via risk-aware costmaps \cite{fanLearningRiskAwareCostmaps2022}, CVaR-based optimization \cite{dixitSTEPStochasticTraversability2024}, chance constraints \cite{freyRoadRunnerLearningTraversability2024}, and speed/clearance policies \cite{benrabahReviewTraversabilityRisk2024}, shifting focus from human-in-the-loop control to autonomous perception and decision-making. Tele-traversability retains a human-centric dimension by encoding operator-specific risk assessments, ensuring autonomous behaviors align with human risk tolerance. Even in high-autonomy scenarios, tele-traversability remains vital for supervisory functions. The system must maintain explainability, particularly during failures, by justifying action selection and clarifying when intervention is required. Thus, tele-traversability serves a \textit{dual} role: enabling human-aligned autonomous navigation while ensuring transparency for effective oversight.

\begin{figure}
  \centering
  \includegraphics[scale=1.2]{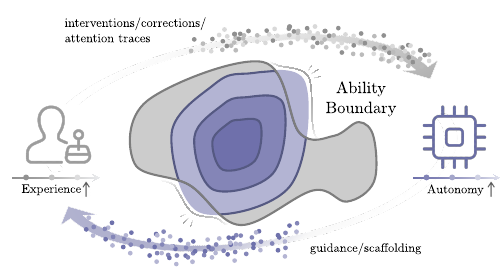}
  \caption{Co-evolution in teleoperation through mutual adaptation.}
  \label{fig_11}
\end{figure}

\subsection{Dynamic Co-evolution of Capability Boundaries}
Teleoperation systems couple human operators constrained by bandwidth-limited sensory interfaces and robotic autonomy operating via local, uncertainty-aware models \cite{darvishTeleoperationHumanoidRobots2023}. Their efficacy depends not on individual agent capabilities alone, but on the dynamic alignment of decision-making competencies under shared environments \cite{musicControlSharingHumanrobot2017, selvaggioAutonomyPhysicalHumanRobot2021}. Building upon the autonomy taxonomy in Section \ref{section:2.3}, we frame teleoperation levels as co-evolving capability boundaries\footnote{The operational regime where an agent can generate decision-theoretic safety and task performance, bounded by perception, computation, and actuation limits.}. This co-adaptation emerges through collaborative learning, context-aware autonomy reconfiguration, and bidirectional competency assessment, finally defining a Pareto-optimal control allocation between human supervision and robotic autonomy.

This coupling follows an evolutionary progression in autonomy. At low autonomy levels, human dominants, with the robot acting primarily as a safeguard actuator while the operator handles high-level reasoning, particularly in unstructured environments where contextual experience is critical \cite{darvishTeleoperationHumanoidRobots2023, selvaggioAutonomyPhysicalHumanRobot2021}. As autonomy advances, the robot's operational boundary expands through improved perception, interaction-aware traversability estimation, and risk-sensitive planning, shifting teleoperation from direct to negotiated control \cite{musicControlSharingHumanrobot2017, luoUsercustomizableSharedControl2024}. This introduces potential human-robot disagreement \cite{gottardiSharedControlRobot2022}. When the robot gains authority to reinterpret or constrain operator inputs, differences in risk assessment escalate from usability issues into safety and performance challenges. Even minor disparities in terrain evaluation can lead to conflicting actions, delayed interventions, or degraded coordination \cite{alonsoSystemTransparencyShared2018}. At near-parallel capability levels, teleoperation becomes cognitive collaboration. The robot contributes structured reasoning (e.g., hazard identification, confidence evaluation) \cite{sakaiExplainableAutonomousRobots2022, setchiExplainableRoboticsHumanRobot2020, schottLiteratureSurveyHow2023}, while the operator provides task intent, contextual knowledge, and preference-based guidance \cite{nikolaidisHumanRobotMutualAdaptation2017}. This dynamic underscores a core design principle for effective human-robot systems: they should collaboratively learn and think \textit{with} humans by maintaining shared representations, ensuring legibility, and adapting to user goals, rather than optimizing solely for robotic performance \cite{sakaiExplainableAutonomousRobots2022, setchiExplainableRoboticsHumanRobot2020, alonsoSystemTransparencyShared2018}.

Thought partnership should not be conflated with consensus in decision-making, as excessive agreement may reflect suboptimal collaboration \cite{nikolaidisHumanRobotMutualAdaptation2017}. An effective teleoperation system must achieve symbiotic capability augmentation, where humans and robots mutually enhance their competencies through bidirectional interactions (Fig. \ref{fig_11}). Robotic autonomy serves as cognitive scaffolding, providing operators with stable traversability cues, predictive trajectory suggestions, and deterministic intervention protocols \cite{luoUsercustomizableSharedControl2024, alonsoSystemTransparencyShared2018}, thereby accelerating skill acquisition through rapid risk calibration. Conversely,  teleoperation supplies robots with rich supervisory signals, such as explicit corrections (e.g., trajectory overrides) and implicit cues (e.g., operational hesitation), enabling continuous refinement of their world models and decision thresholds \cite{mandlekarHumanintheLoopImitationLearning2020, liuRobotLearningJob2025}. This framework aligns with cognitive science principles; that is, human-level robustness arises from structured knowledge (e.g., causal graphs, intuitive priors, and compositional representations) rather than purely pattern recognition\cite{lakeBuildingMachinesThat2017, collinsBuildingMachinesThat2024}. Teleoperation uniquely instantiates this by jointly training on sensorimotor correlations and interactive protocols that encode human reasoning, correction, and adaptation \cite{mandlekarHumanintheLoopImitationLearning2020, liuRobotLearningJob2025}. By explicitly modeling both the world and the operator, the system enables true thinking and collaboration \textit{with} humans. Thus, tele-traversability should embody an adaptive shared representation by quantifying reliability, transparently justifying autonomy interventions, and closing the adaptation loop via continuous feedback to update traversability criteria and their interpretable presentation.

\section{Traversability From Autonomy to Teleoperation: Methods and Gaps} 
\label{section:3}
Current traversability research has largely focused on autonomous ground robots, employing multi-sensor perception and risk modeling for motion planning. While these methods remain applicable for teleoperated robots with high autonomy, teleoperation introduces a critical factor: the human operator. This shift necessitates integrating human factors (e.g., subjective perception, intent alignment) and control dynamics (e.g., shared autonomy, interactive feedback) alongside physical feasibility. Such an expanded framework better reflects real-world human-robot collaboration, bridging physical modeling with human-centric interaction.

\subsection{Evolution of Tele-Traversability Representation}
\label{section:3.1}
\begin{figure}[t]
  \centering
  \includegraphics[scale=1.2]{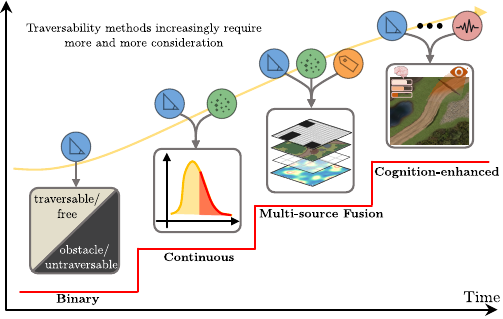}
  \caption{Evolution of environmental representations for traversability assessment in off-road teleoperation.}
  \label{fig_7}
\end{figure}

Traversability representation has evolved from binary occupancy maps to continuous cost fields and multi-layer semantic risk maps with uncertainty (Fig. \ref{fig_7}). Early works adopted binary or discrete grid cells  \cite{serajiFuzzyTraversabilityIndex2000} mapped terrain features to qualitative levels (e.g., good/poor) for navigation \cite{serajiBehaviorbasedRobotNavigation2002} or safe/unsafe thresholds based on terrain geometry \cite{genneryTraversabilityAnalysisPath1999}. Howard \cite{howardQuantifyingTraversabilityTerrain2005} formalized traversability using terrain attributes, while vision-based methods produced binary (traversable/non-traversable) maps from appearance features  \cite{howardVisionbasedTerrainCharacterization2001}. Probabilistic mapping approaches like relative probabilistic mapping \cite{chenRealtimeRelativeProbabilistic2015} discretized continuous measurements (e.g., height, point densities) into three classes (traversable/obstacles/unknown) for planning. While efficient for real-time teleoperation, such discrete representations lack the granularity needed for complex autonomous decision-making.

Later research shifted from discrete labels to continuous traversability metrics, better capturing navigation challenges. Early terramechanics approaches \cite{iagnemmaTerrainEstimationHighspeed2002, iagnemmaOnlineTerrainParameter2004} estimated real-time terrain parameters (e.g., stiffness, sinkage) from vehicle dynamics, enabling high-speed traversability prediction via explicit vehicle-terrain interaction models. For planetary rovers, Ishigami et al. \cite{ishigami2006terramechanics, ishigamiTerramechanicsbasedModelSteering2007} introduced contact mechanics-based continuous metrics for slope traversability and loose-soil steering, eliminating binary thresholds. Learning-based methods \cite{hoNeartofarNonparametricLearning2013, hoTraversabilityEstimationPlanetary2013} further advanced this paradigm by inferring traversability from near-field sensors (stereo vision, proprioception) using non-parametric models, handling occluded and deformable terrains. Modern frameworks  \cite{serajiNewTraversabilityIndices2003, grafOptimizationBasedTerrainAnalysis2019} unified these metrics with path planning via joint-terrain cost formulations. As surveys \cite{borgesSurveyTerrainTraversability2022,beycimenComprehensiveSurveyUnmanned2023} indicate, current systems fuse geometric and proprioceptive data into continuous cost maps as standard inputs for motion planners. These representations enhance teleoperation via intuitive visualizations (e.g., risk heatmaps) while remaining real-time performance.

Recent advances in traversability representation now encode semantics, uncertainty, and risk metrics alongside geometric costs. Modern methods enrich grid maps with multi-modal data, including geometric properties (e.g., elevation/slope), semantic labels (e.g., terrain types), and fused sensor attributes (LiDAR, cameras, etc) \cite{hosseinpoorTraversabilityAnalysisSemantic2021, mengTerrainNetVisualModeling2023, huSurveyMultisensorFusion2020}. For instance, TerrainNet \cite{mengTerrainNetVisualModeling2023} unifies geometric and semantic data in gravity-aligned grid structures. Surveys \cite{sevastopoulosSurveyTraversabilityEstimation2022, borgesSurveyTerrainTraversability2022, beycimenComprehensiveSurveyUnmanned2023} highlight the integration of confidence metrics (e.g., variance, sensor disagreement) per cell, capturing both terrain appearance and perception reliability. Such granular representation enables risk-aware navigation. For instance, the STEP framework models traversability cost as random variables, aggregated via coherent risk measures (e.g., CVaR) to generate risk maps with visual encodings for expected difficulty and tail risks \cite{dixitSTEPStochasticTraversability2024, akellaSamplebasedBoundsCoherent2024}. Similarly, self-reflective terrain adaptation and covariance-steering model predictive control methods integrate probabilistic safety constraints into decision-making \cite{sivaSelfReflectiveTerrainAwareRobot2021, knaupSafeHighPerformanceAutonomous2023}. For teleoperation, these maps provide human-interpretable visualizations (e.g., color-coded risk gradient, annotated semantics for loose gravel, and explicit uncertain regions), enhancing operator situational awareness and trust in autonomy.

In teleoperation, environment representations serve as planning tools for robots while communicating information to human operators. Prior work identifies three challenges for feedback systems: unpredictable delays, bandwidth limitations, and the need to maintain operator situational awareness \cite{chopraBilateralTeleoperationInternet2003, hokayemBilateralTeleoperationHistorical2006}. Research shows that multi-modal feedback (e.g., videos, annotated maps) helps operators rapidly assess environmental layout, robotic capabilities, and hazards \cite{goodrichHumanRobotInteraction2008, goodrichTeleoperationAssistiveHumanoid2013, debarrosEnhancingRobotTeleoperator2011}. Current representation methods fall into three categories:
\begin{itemize}
    \item \textbf{Binary masks} (traversable/non-traversable areas) are bandwidth-efficient and easily overlaid on video, but lack granularity and risk quantification.
    \item \textbf{Continuous cost maps} offer richer information but risk overwhelming operators without careful design.
    \item \textbf{Semantic-uncertainty layers} combine intuitive visual cues (e.g., color-coded risk zones) with structured haptic feedback to balance clarity and precision \cite{debarrosEnhancingRobotTeleoperator2011}.
\end{itemize} 

\subsection{Environmental Perception and Terrain Modeling}
\label{section:3.2}
Environmental perception and terrain modeling are the basis of joint traversability estimation. Robots and human operators first require (direct/indirect) environment representations derived from sensory data. For human-robot collaboration, these representations must incorporate human-understandable factors and transparently communicate the robot's decision-making rationale to operators.

\subsubsection{Obstacle and hazard cues} Obstacle detection is critical for terrain classification and traversability analysis \cite{borgesSurveyTerrainTraversability2022}. Traditional systems conservatively mark no-go regions, but tele-traversability demands more nuanced interpretation. For example, geometry-based methods may misclassify traversable features (e.g., bushes) as obstacles due to height discontinuities \cite{dubbelmanObstacleDetectionDay2007, mortonPositiveNegativeObstacle2011}. Recent approaches to this by incorporating semantic information through visual \cite{xieMonocularVisionBased2017, shabanSemanticTerrainClassification2022} or multimodal sensing \cite{liuPositiveNegativeObstacles2024}. These methods encode human priors about object-traversability relationships, shifting the challenge from detection to contextual interpretation. Thus, operator expertise becomes crucial; human excel at judging conditional hazards and suggesting negotiated strategies \cite{fongMultirobotRemoteDriving2003, darvishTeleoperationHumanoidRobots2023}. Multimodal fusion enhances this process by combining geometric, semantic, and other features \cite{schillingGeometricVisualTerrain2017, reedAutonomousHikingTrail2025}, but final traversability decisions (i.e., whether pre-programmed or interactive) ultimately benefit from human judgment.

Existing methods typically classify terrain into free, obstructed, and unknown regions \cite{kuthirummalGraphTraversalBased2011}, imposing hard or near-hard mobility constraints by assigning prohibitive crossability costs \cite{molinoTraversabilityMetricsRough2007}. While effective for identifying macroscopic hazards (e.g., high steps or gaps), such binary classifications neglect graded terrain difficulty and fail to explain why certain regions are traversable but risky. Instead of rigidly labeling all hazards as obstacles, the system should maintain provisional classification for ambiguous regions and refine them using either expert knowledge \textit{or} real-time operator input. This approach is particularly valuable since hazard indicators, compared to detailed terrain analysis, provide operators with immediate, legible traversability information. For instance, visual cues (e.g., warning zones and highlighted boundaries) can effectively convey motion constraints, enhancing the operator's understanding of robot behavior \cite{nielsenEcologicalInterfacesImproving2007, kamezakiSituationalUnderstandingEnhancer2021}.

\subsubsection{Geometric terrain features}
\label{section:3.2.2} 
Geometric terrain features, extracted from LiDAR or RGB-D sensors, quantify terrain properties using grid-based representations \cite{ruetzOVPCMesh3D2019, fankhauserProbabilisticTerrainMapping2018}. These features transform raw sensor data into standardized metrics (e.g., elevation and curvature) for downstream modules such as obstacle detection. Unlike binary obstacle maps, which simply label traversability, geometric features provide continuous descriptors of terrain difficulty for nuanced path planning. 

Geometric cues are the backbone of traversability assessment due to their computational efficiency and physical interpretability. These methods quantify terrain properties using elevation-derived metrics such as height variation, slope, and roughness \cite{reina3DTraversabilityAwareness2014, schillingGeometricVisualTerrain2017}. Early work \cite{langerBehaviorbasedSystemOffroad1994, genneryTraversabilityAnalysisPath1999} demonstrated that basic height differentials could identify navigable regions, while modern approaches combine multi-dimensional geometric features (e.g., slope-step-roughness triplets \cite{beycimenRealTimeTerrainTraversability}) to assess terrain stability and robot interaction potential \cite{belloneNewApproachTerrain2013, guanTNESTerrainTraversability2023}.  However, geometric measurements alone are seldom sufficient for traversability assessment, as they lack semantic meaning for both operators and robots. The challenge lies in aggregating these measurements into interpretable terrain structures. Here, human expertise plays a pivotal role by contextualizing how geometric data translates to traversal difficulty for a specific robot and task. For instance, a step height may be objectively measured, but its interpretation (e.g., negligible, manageable, or prohibitive)  depends on the robot's mobility constraints. Operator expertise becomes crucial for evaluating control-intensive maneuvers, where subjective tolerance thresholds vary. Most autonomous systems implicitly embed this \textit{geometric-to-semantic} mapping  \cite{oliveiraThreeDimensionalMappingAugmented2021, e.carvalho3DTraversabilityAnalysis2024}, either through hand-crafted rules or annotated training labels. \textit{Tele-traversability formalizes this process}, aligning robot decisions with operator expectations, while preserving objectivity. These mappings serve as internal representations for autonomy and provide explainable cues (e.g., speed reduction or re-planning triggers) when visualized through physically meaningful interfaces \cite{kamezakiSituationalUnderstandingEnhancer2021, chaeDivergentEffectsVisual2024}. 


\subsubsection{Appearance and semantic features.} 
Appearance and semantic features encode terrain types and material properties to improve traversability assessment. While geometry captures physical terrain morphology, appearance and semantics reveal surface identity, exposing mobility hazards (e.g., mud, dense vegetation) that may be geometrically subtle but functionally critical \cite{angelovaLearningPredictionSlip2007, procopioLearningTerrainSegmentation2009}. These methods typically extract image-based descriptors (e.g., color, texture) to classify surfaces into predefined semantic categories (e.g., road, mud). However, this characterization relies on human expert priors, as both category definitions and granularity are task-dependent and designer-specified. By configuring these semantic representations, operators enhance system interpretability, aligning robot decision-making with human intuition.

Early vision-based traversability estimation methods relied on low-level visual cues (e.g., color, texture) for terrain segmentation and traversability inference \cite{manduchiObstacleDetectionTerrain2005, angelovaLearningPredictionSlip2007}. Modern approaches retain this paradigm but leverage learned visual representations instead of hand-crafted features, enabling end-to-end traversability prediction from raw RGB inputs \cite{jungVSTRONGVisualSelfSupervised2024, vecchioTerrainTraversabilityPrediction2024, mattamalaWildVisualNavigation2025}. A key advancement is semantic terrain categorization, where traversability costs are derived from labeled regions \cite{hosseinpoorTraversabilityAnalysisSemantic2021, leeTerrainawarePathPlanning2025}, improving interpretability by explicitly linking surface semantics to traversal risks. Further refinements integrate semantic reasoning with geometric representations through multimodal fusion, for instance, by combining LiDAR-based semantic segmentation with elevation mapping \cite{miliotoRangeNet++Fast2019, cortinhalSalsaNextFastUncertaintyAware2020}, or jointly optimizing semantics and geometric consistency \cite{mengTerrainNetVisualModeling2023, sivaSelfReflectiveTerrainAwareRobot2021}. This integration improves prediction accuracy and generates a human-legible explanation, aligning robotic perception with operator intuition. 

The advent of foundation models has revolutionized semantic representation in robotics. Unlike fixed taxonomies, promptable vision-language models (e.g., SAM3 \cite{carionSAM3Segment2025}) now enable terrain parsing using human-interpretable concepts. Transformer-based backbones like SegFormer provide robust feature extraction \cite{xieSegFormerSimpleEfficient2021}, while Segment Anything models support open-vocabulary segmentation via natural language prompts or exemplar images \cite{kirillovSegmentAnything2023, raviSAM2Segment2024}. This eliminates task-specific networks, directly mapping descriptors (e.g., deep mud, standing water, or tall vegetation) to semantic risk layers. Thus, traversability estimation becomes more flexible, requiring minimal annotation effort even for rare terrain categories.

\begin{figure}
  \centering
  \includegraphics[scale=1.0]{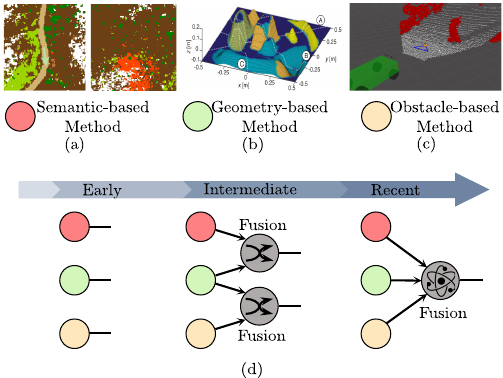}
  \caption{Evolution of front-end representations for off-road traversability from isolated cues to complementary fusion. (a-c) Representative pipelines built on semantic- \cite{mengTerrainNetVisualModeling2023}, geometry- \cite{fankhauserProbabilisticTerrainMapping2018}, and obstacle-based \cite{aeschimannGroundObstaclesDetecting2015} evidence, respectively. (d) Historical trend: early methods typically rely on a single cue, intermediate approaches perform pairwise fusion, and recent systems increasingly integrate all three sources in a unified fusion stage.}
  \label{fig_6}
\end{figure}

Traversability assessment should integrate complementary information layers (Fig. \ref{fig_6}). Obstacle layers define safety boundaries, geometric layers quantify terrain support and mobility constraints, and semantic layers capture material/contextual risks undetectable through geometric alone. This shift toward unified multi-layer mapping emphasizes the fusion of heterogeneous cues into a cohesive representation \cite{beycimenRealTimeTerrainTraversability, zhangAutonomousVehiclesTraversability2024, zhouTerrainTraversabilityMapping2022}. Modern systems extend this framework by incorporating robot-centric factors, such as proprioceptive feedback, interaction outcomes, and mission-specific costs, thereby enabling dynamic traversability evaluation \cite{ederTraversabilityAnalysisOffroad2023, castroHowDoesIt2023}. Such representations align with human operators' cognitive processes, jointly assessing environmental features, traversal consequences, and confidence levels.

\subsection{Robot Modeling and Robot-Terrain Interaction}
For teleoperated ground robots, traversability assessment should consider robot-terrain interactions \cite{shuOverviewTerrainTraversability2025}, such as contact force distribution, slip dynamics, and failure modes. 

\subsubsection{Morphology and kinematic-geometric constraints}

Morphology determines traversability by defining how a robot interacts with terrain and the kinematic-geometric constraints that govern feasible motion. Traversability is not a sole property of terrain but rather a platform-dependent relationship. In other words, elevation changes, discontinuities, or narrow passages only become meaningful relative to the robot's motion envelope. Thus, traversability varies across different platforms, even in identical environments.

For wheeled and tracked robots, traversability depends on maintaining continuous ground contact while satisfying clearance and steering constraints. However, slip, attitude variations, and terrain-induced body motion further bound mobility by affecting controllability and stability on uneven terrain. Reina and Galati \cite{reinaSlipbasedTerrainEstimation2016, reinaTerrainAssessmentPrecision2017}, for instance, demonstrated that terrain-induced motion states (e.g., slip and inertial responses) are intrinsic to traversability limits rather than secondary effects. Subsequent planning incorporated suspension kinematics or reduced-order vehicle dynamics to predict chassis response to terrain \cite{zhangTraversabilityAssessmentTrajectory2019, gonzalezRobustTubebasedPredictive2011, blackmoreChanceConstrainedOptimalPath2011}.

Legged systems (e.g., quadrupeds and humanoids) operate under fundamentally different constraints than wheeled or tracked robots, as traversability depends on feasible foothold selection, posture stability, and dynamically consistent support transitions under whole-body balance and joint limits. Loc, et al. \cite{locImprovingTraversabilityQuadruped2011} showed that body configuration directly affects foothold viability, meaning traversability is determined by the coupling between robot and terrain. Later works formalized this insight through whole-body optimization frameworks that integrate support conditions, posture feasibility, and balance constraints \cite{liRuggedTerrainTraversability2019, hutterANYmalHighlyMobile2016, hoellerANYmalParkourLearning2024, goswami2019humanoid}. Thus, a given terrain feature (e.g., a gap or cluttered passage) may be traversable for one robot yet impassable for another. 

\subsubsection{Robot-terrain interactions}
Conventional geometric representations or dynamic models often fail to capture the state-dependent coupling between terrain properties and robotic mobility (Section \ref{section:3.2}). Critical mobility constraints (e.g., slope slip, soft-terrain sinkage, and rough-terrain disturbance) necessitate traversability assessment frameworks that explicitly model interaction-derived risks (e.g., slip, collision) during planning \cite{angelovaLearningPredictionSlip2007, reinaTerrainAssessmentPrecision2017}. This paradigm extends to tele-traversability, where conveying physical interactions enhances operator situational awareness. Field operators may overlook subtle interaction cues (e.g., incipient slip via chassis vibrations) detectable by onboard sensors. Recent telepresence solutions address this by multisensory feedback. For instance, Huang et al. \cite{huangTelepresenceAugmentationVisual2024} reduced workload via visual-haptic rendering, while Dafarra, et al. \cite{dafarraICub3AvatarSystem2024} enabled full-body haptic embodiment, including weight and touch feedback. These advances demonstrate that transmitting physical interactions, beyond visual data, is critical for accurate terrain assessment and risk anticipation.

Autonomous systems utilize interaction dynamics to enhance predictive mobility assessment. For instance, Ugenti et al. \cite{ugentiLearningPredictionVehicleterrain2021} established preemptive motion resistance estimation using standoff terrain observations. However, as Fu, et al. \cite{fuAnyNavVisualNeuroSymbolic2025} argue, generalizable mobility prediction requires neuro-symbolic reasoning to map visual terrain features to physical-based interaction parameters. Xu et al. \cite{xuVertiBenchGeneralScalable2025} further emphasize the necessity of dynamic interaction evaluation during execution, as static perception cannot anticipate action-dependent failure modes. These works underscore two key research priorities: (i) advancing onboard, interaction-aware traversability prediction, and (ii) developing effective operator feedback mechanisms for shared autonomy systems.

\section{Future Directions: Interaction-based, Cognition-informed, and Autonomy-level-aware}
Tele-traversability evolves from a robot-centric perception problem to a cognitive coupling challenge when human operators are involved. In teleoperation, traversability estimation simultaneously serves two roles (Section \ref{section:2.2}): 
\begin{itemize}
    \item \textbf{Adaptive feasibility}: Beyond geometric constraints, traversability assessments should incorporate operator factors (e.g., risk tolerance, control preference) to bridge the gap between robotic and human evaluations. 
    \item \textbf{Interpretable interface}: The representation must provide intuitive alignment through abstraction and visualization, allowing operators to rapidly comprehend the robot behavior.
\end{itemize}
This duality suggests two critical thrusts:
\begin{itemize}
    \item \textbf{Dynamic interaction modeling}: Traversability should be reframed as an emergent property of the human-robot-environment triad.
    \item \textbf{Cognition state integration}: Real-time modeling of operator factors (e.g., attention allocation, workload, situational awareness, trust dynamics, and risk perception) is essential to maintain human-robot evaluation consistency during mission execution.
\end{itemize}
Notably, these approaches must be \textit{autonomy-level-adaptive}, as traversability functionally morphs across operational modes, for instance, providing visual cues for manual control, negotiation parameters for shared autonomy, and policy boundaries for autonomous operation.

\subsection{From Terrain Property to Human-Robot-Environment Interaction: Embodied Robotics}

Tele-traversability fundamentally aligns with embodied intelligence principles, where intelligent behavior emerges from continuous agent-environment interaction rather than abstract reasoning alone \cite{brooksIntelligenceRepresentation1991, pfeiferSelfOrganizationEmbodimentBiologically2007}. Teleoperation extends this paradigm by establishing a distributed control loop comprising the human operator, the robotic agent, and the environment. This system exhibits dual interaction processes: (i) robot-environment interaction governing physical traversal feasibility, and (ii) human-robot interaction mediating perception and control \cite{toetEnhancedTeleoperationEmbodiment2020}.

\begin{figure}[t]
  \centering
  \includegraphics[width=\linewidth,height=0.60\textheight,keepaspectratio]{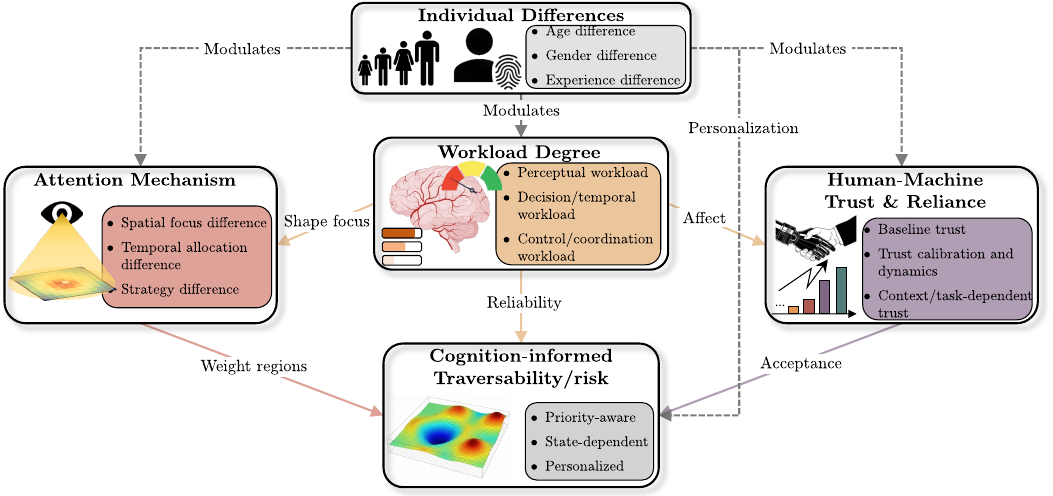}
  \caption{Cognition-informed tele-traversability. Attention mechanisms weight terrain regions according to the operator’s focus, workload degree affects the reliability and conservatism of human judgment, and human-machine trust/reliance governs acceptance and use of autonomy outputs. Individual differences (e.g., experience and age) modulate these cognitive factors and enable personalization. Together, these influences shape a priority-aware, state-dependent, and personalized traversability/risk representation that better aligns machine assistance with the operator’s subjective assessment.}
  \label{fig_9}
\end{figure}

Unlike autonomous navigation that evaluates traversability as static terrain properties, tele-traversability emerges dynamically from coupled human-robot perception and action. Robots assess terrain through onboard sensors under uncertainty, while operators interpret environmental feedback through interfaces. This dichotomy can create perceptual mismatches even when individual estimates are locally valid. 

Recent works on embodied robotics underscore interaction centrality in teleoperated systems. Some platforms leverage teleoperation for demonstration collection in learning-based control \cite{posadas-navaBEAVRBimanualMultiEmbodiment2025}, where interface design and embodiment choices critically impact operator performance (e.g., workload, task efficiency) and data quality \cite{moyenRoleEmbodimentIntuitive2025}. Advances in compliant anthropomorphic designs further enable natural human-robot physical interaction in contact-rich tasks \cite{jungeADAPTTeleopRoboticHand2025}. These developments necessitate rethinking tele-traversability beyond geometric terrain analysis. Successful teleoperated navigation depends on (i) physical dynamics at robot-terrain interaction (e.g., contact, terrain deformation), (ii) cognitive processes in human-robot communication (e.g., state perception, decision-making). Future tele-traversability representations must therefore jointly model robot-terrain physical interaction constraints and cognitive alignment between operator intent and robotic capability.

\subsection{From Objective Estimation to Cognitive Alignment} 
\label{section:4.2}
Onboard intelligence advancements can induce human-robot discrepancies in terrain traversability assessment, creating control conflicts \cite{methnaniWhosChargeHere2024, sheridan1992telerobotics}. These mismatches stem from two sources. The first one is the gap between subjective human judgment and algorithmic traversability quantification, exacerbated by teleoperation's situational awareness constraints \cite{endsleyTheorySituationAwareness2011}. The second one is the latency-induced performance degradation that escalates the operational cognitive workload \cite{chenEffectsCommunicationDelay2025, kamtamNetworkLatencyTeleoperation2024}. Addressing latency requires joint consideration of machine control and human adaptation\cite{duSensoryManipulationCountermeasure2024}, and autonomy levels impact the operator's cognitive load and trust \cite{panEffectsSharedControl2024}.

Human's traversability assessment is cognition-bound, mediated by indirect perception, selective attention, and situational awareness. Unlike autonomous systems that perform continuous environment evaluation, operators must infer terrain properties through delayed, bandwidth-limited feedback channels  \cite{endsleyTheorySituationAwareness2011}. This forces reliance on heuristic strategies with selective attention to salient regions rather than comprehensive coverage \cite{kamtamNetworkLatencyTeleoperation2024, fong2001collaboration}. Teleoperation awareness is also task-dependent and can not be resolved through mere information augmentation \cite{duSensoryManipulationCountermeasure2024}. The resultant perception is intrinsically local and subjective, shaped by the operator's attentional focus, cognitive workload, and trust in the robot (Fig. \ref{fig_9}). This contrasts significantly with the algorithmic estimates' global, continuous nature. These key mechanisms amplify the divergences of \textit{attention allocation}, \cite{draheimRoleAttentionControl2022}, \textit{cognitive workload} \cite{longoHumanMentalWorkload2022}, 
\textit{human-robot trust} \cite{matthewsIndividualDifferencesTrust2020}. By adapting robot intelligence to individual operator characteristics (e.g., competence level and behavioral tendencies), we can achieve consistent collaboration across diverse human operators while maintaining seamless control transition.

\emph{Attention allocation}. As established in Section \ref{section:2.2}, tele-traversability emerges from the interplay of terrain properties, robot mobility constraints, and the operator's acceptability. This acceptability depends not on all objectively available information, but rather the subset selectively attended to and processed under cognitive constraints \cite{posnerOrientingAttention1980}. This distinction between \emph{information availability} and \emph{information uptake} is fundamental to teleoperation, where operators must integrate heterogeneous data streams (e.g., terrain maps, camera feeds, robot states) while controlling the platform simultaneously. Attention allocation serves as the cognitive gateway for terrain-relevant information to enter operator assessment. Rather than being a peripheral human factor, it structures tele-traversability judgment through:
\begin{itemize}
    \item Spatial scoping: Defining what terrain regions, hazards, or path segments are evaluated at a given moment, while others remain marginalized \cite{rangelovEvidenceAccumulationPerceptual2020, karpinskyAutomationTrustAttention2018}.
    \item Evidential weighting: Amplifying salient cues (e.g., perceived hazards, state changes) while discounting less attended features, regardless of objective importance \cite{karpinskyAutomationTrustAttention2018}.
    \item Temporal dynamics: Governing when judgments are updated as the robot moves and new information arrives \cite{kamezakiSituationalUnderstandingEnhancer2021, chaeDivergentEffectsVisual2024}.
\end{itemize}
Thus, attention allocation functions more than filter inputs, but structures how and when traversability judgments are updated.

Attention should not be modeled merely as a gaze trace but as a latent state reflecting how operators prioritize terrain- and risk-relevant information. Selective attention shapes perceptual decision-making by modulating the weight of sensory inputs  \cite{rangelovEvidenceAccumulationPerceptual2020}. Future models must estimate both attention focus and which critical terrain cues were likely integrated or missed due to competing task demands \cite{karpinskyAutomationTrustAttention2018}. Such limitations manifest in distinct tele-traversability errors. \textit{Divided attention} fragments assessment across tasks, \textit{inattentional blindness} excludes visible hazards from judgment, and \textit{looked-but-failed-to-see} failures yield superficial visual coverage without reliable recognition \cite{wolfeNormalBlindnessWhen2022}. These challenges intensify when operators must combine map-level context with local camera feeds, as attention to one often compromises integration of the other \cite{kamezakiSituationalUnderstandingEnhancer2021, belardinelliGazeBasedIntentionEstimation2024, anOperatorVisualAttention2024}. Attention-aware tele-traversability models should identify gaps in operator-side evidence to explain human-robot disagreement and trigger adaptive safeguards when terrain risks are overlooked.


\emph{Cognitive workload}. In teleoperation, the critical challenge lies not in the sheer volume of information but in the operator's capacity to synthesize it into stable, well-calibrated traversability judgments \cite{carissoliMentalWorkloadHumanRobot2024,duSensoryManipulationCountermeasure2024,longoHumanMentalWorkload2022}. As cognitive load increases, judgments become less integrative and more reliant on local heuristics. 

Cognitive workload influences tele-traversability through two primary pathways (Fig.~\ref{fig_9}): A \textit{direct} effect on judgment formation and \textit{indirect} effects mediated by attention allocation and human-machine trust. The \textit{direct} effect manifests in the operator's diminished ability to assess map cues, video feedback, robot state, and uncertainty. Instead, they increasingly depend on the most immediate or salient cues, leading to less stable and poorly calibrated judgment, neither consistently conservative nor aggressive but rather inconsistently weighted. The \textit{indirect} effects arise as workload narrows attentional scope, causing operators to focus on dominant visual or control channels while overlooking peripheral hazards or uncertainty cues. In addition, high workload reduces the capacity to verify autonomy outputs, fostering greater reliance on automation while making trust more brittle when robot behavior proves ambiguous \cite{anOperatorVisualAttention2024,liuOperationalPerformanceCognitive2025}. Shared-control studies further indicate that autonomy design jointly shapes workload and trust, rather than affecting them independently \cite{devisserTheoryLongitudinalTrust2020,panEffectsSharedControl2024}.

Tele-traversability necessitates modeling operator workload not just for post hoc evaluation but also for real-time adaptation of representation and assistance. Workload can be estimated from subjective measures (e.g., NASA-TLX, Bedford scales), secondary-task methods assessing attentional capacity, or multimodal indicators such as task performance, control inputs, eye tracking, and physiological signals \cite{hartDevelopmentNASATLX1988, casaliComparisonRatingScale1983, koschSurveyMeasuringCognitive2023, odohPerformanceMetricsTeleoperation2024}. These estimates enable adaptive interfaces; for instance, suppressing nonessential terrain details while highlighting critical regions or enforcing stricter confirmation for high-risk actions when operator judgment is compromised \cite{linPerceptionActionAugmentation2024, panEffectsSharedControl2024}. Thus, workload modeling ensures alignment between human decision-making and robotic traversability reasoning during challenging remote navigation.

\emph{Human-machine trust}. While cognitive workload affects the reliability of operator's assessments, trust determines whether the operator actually incorporates the robot's tele-traversability information into decision-making. In teleoperation, trust is not as a static attitude toward the robot but rather a dynamic willingness to rely on its traversability perception, uncertainty estimates, and autonomous navigation decisions. Like broader trust-in-automation research, trust here is multidimensional and time-varying, shaped by perceived reliability, competence, transparency, predictability, and safety, rather than reducible to a single fixed metric \cite{hoffTrustAutomationIntegrating2015,kohnMeasurementTrustAutomation2021,yangQuantifyingTrustDynamics2023}. Trust in tele-traversability function-specific; that is, an operator might trust the robot's terrain mapping but distrust its path planning or speed adjustments. The key question is not just whether the robot \textit{can} estimate traversability but whether it \textit{effectively communicates} its limits and intervenes in alignment with operator intent.

Well-calibrated trust enables operators to integrate robot-generated warnings, uncertainty displays, slowdowns, and reroutes into terrain navigation, reducing conflicts and improving coordination. Conversely, distrust leads to autonomy disuse, frequent overrides, and excessive monitoring, while overtrust risks uncritical reliance on flawed autonomy, especially when perception is uncertain or terrain exceeds robot capabilities \cite{devisserTheoryLongitudinalTrust2020,methnaniWhosChargeHere2024,panEffectsSharedControl2024}. Recent work highlights that reliance depends on how clearly the robot exposes its reasoning, confidence, and operational limits. Explainable and adaptive robot behavior enhance perceived trustworthiness \cite{cantucciTrustworthinessAssessment2025}, while delayed teleoperation increasingly relies on transparency-aware arbitration to stabilize human-autonomy reliance \cite{gulecyuzEnhancingSharedAutonomy2025}. 

Trust bridges machine-side feasibility and operator-side acceptability, warranting treatment as a dynamic latent state, not just a static survey measure. Current assessment methods include self-reports, behavioral indicators (e.g., override frequency, compliance with recommendations), monitoring patterns, eye movement, and physiological signals, with growing emphasis on multimodal fusion for real-time estimation \cite{mengInterpretableTrustAssessment2026, rindfussModelingTrustDynamics2025}. These observations can then be embedded in dynamic models, such as Bayesian estimators, state-space models, and learned temporal predictors, to track trust evolution and enable adaptive calibration \cite{yangQuantifyingTrustDynamics2023,guoModelingPredictingTrust2021}. For tele-traversability, such models should be function-aware and context-aware, separately estimating trust in terrain perception, uncertainty visualization, shared control, and intervention policies. 

\emph{Individual differences}. Research on driving behavior shows systematic individual variations in risk preference, following and braking strategies, aggressiveness, and style adaptation, which influence how uncertain situations are interpreted and when corrective actions are taken \cite{wangReviewDriverBehavior2014,zhangShareablePotentialDriving2024}. Operator risk judgments cannot be treated as a fixed, uniform threshold. The same terrain cue, uncertainty estimate, or robot action may be acceptable to one operator but rejected by another, depending on experience, behavioral tendencies, and risk tolerance. In fact, human judgment is not a direct readout of objective hazard but is shaped by subjective weighting and systematic deviations from normative rational models \cite{slovicPerceptionRisk2000, ProspectTheoryAnalysis}. In teleoperation, these differences are further amplified by indirect perception, delayed feedback, and varying levels of trust in robot autonomy \cite{devisserTheoryLongitudinalTrust2020}. Rather than merely introducing noise, individual differences fundamentally shape how traversability information is interpreted and acted upon.

Tele-traversability models must account for inter-individual variability in subjective risk perception, intervention tendencies, and acceptance thresholds, rather than assuming a universal mapping from robot-side feasibility to human-side acceptability. This underscores the need for adaptive explanation, transparency, and authority allocation mechanisms that align robot behavior with operators' diverse experience, trust disposition, and risk tolerance \cite{methnaniWhosChargeHere2024,verhagenMeaningfulHumanControl2024}. 

\section{Open Challenges and Perspectives}
\label{section:5}
Teleoperation remains critical for deploying robots in unstructured or hazardous environments where full autonomy is unreliable. From planetary exploration and disaster response, these systems rely on human operators to handle uncertainty, sparse sensing, and unforeseen terrain conditions. Scalable deployments confirm that human supervision is indispensable under high-risk, high-uncertainty scenarios \cite{darvishTeleoperationHumanoidRobots2023}. While terrain traversability estimation is widely recognized as essential for safe navigation in unstructured environments \cite{borgesSurveyTerrainTraversability2022, beycimenRealTimeTerrainTraversability}, its practical impact on teleoperation systems remains limited. Most deployed platforms still utilize simplified heuristics or basic geometric representations. This gap between research and deployment arises from three key challenges: 
\begin{itemize}
\item [1)] \textbf{Operator-centric modeling}: Current traversability models neglect the operator's perception, personalization, and cognitive state. 
\item [2)] \textbf{Intuitive representation}: Existing methods fail to communicate the robot's reasoning in an intuitive and predictable manner.
\item [3)] \textbf{Human-factor integration}: While this work proposes a traversability model incorporating human factors, further refinement is needed.
\end{itemize}

As discussed in Section \ref{section:3}, most existing approaches model terrain through geometric or semantic properties, generating cost maps or passability scores that are implicitly treated as objective. While suitable for autonomous planners, these representations fail to capture how humans evaluate terrain difficulty during teleoperation. Empirical studies in human-robot interaction show that operator behavior, decision thresholds, and intervention strategies vary with cognitive workload, situational awareness, and perceived task difficulty. As a result, the same terrain may be deemed acceptable or unacceptable under different operational contexts, even when its physical properties are unchanged. Treating traversability as a static terrain attribute thus overlooks its inherently contextual nature. Instead, traversability should be viewed as a property emerging from the interplay between terrain, robot state, task demands, and human judgment \cite{saucedoEATEnvironmentAgnostic2024}.

Recent advances in probabilistic and risk-aware traversability estimation partially address this limitation by moving beyond binary or deterministic metrics. However, even when uncertainty is modeled, a gap persists between \textit{mathematically defined risk} and \textit{operator-perceived risk}. Risk-theoretic robotics highlights that the choice of risk metric is not neutral; it encodes normative assumptions about undesirable outcomes and tail-event weighting, directly shaping behavior under uncertainty \cite{majumdarHowShouldRobot2020}. In teleoperation, this distinction is critical because operators must intuitively understand and calibrate risk-aware behavior rather than treating it as an opaque algorithmic output. 

Furthermore, research on trust and automation emphasizes that human reliance on autonomy depends on how reliability is communicated and experienced over time. Miscalibrated trust --- whether due to overreliance or excessive intervention --- can degrade performance, even when the underlying system is robust \cite{kohnMeasurementTrustAutomation2021, yangQuantifyingTrustDynamics2023}. Thus, traversability representations that encode risk without conveying its implications to the operator may risk undermining trust and reducing the practical value of risk-aware navigation.

Teleoperation traversability must serve as a bridge between robot and operator, translating terrain difficulty and anticipated consequences into an intuitive form. Beyond merely guiding planning, it should help operators anticipate robot behavior, interpret action recommendations, and adjust their decisions accordingly. This aligns with core principles of human-robot interaction --- like legibility and predictability --- where motion and decisions must balance optimality with interpretability \cite{draganLegibilityPredictabilityRobot2013}. Similarly, explanation research frames ``robot reasoning disclosure” as a model alignment problem; that is, explanations should reconcile discrepancies between the robot's logic and the operator's mental model to ensure understandable and acceptable behavior \cite{chakrabortiPlanExplanationsModel2020}. Thus, traversability should not only inform onboard intelligence but also act as an operator-facing medium to reduce ambiguity and support robust decision-making.

These insights reshape how traversability methods are evaluated. While conventional metrics (e.g., classification accuracy, regression error) remain essential for perceptual fidelity, they are inadequate once traversability becomes a teleoperation-oriented representation. As discussed in Section \ref{section:2.2}, its value now hinges not just on prediction correctness but also on operator acceptance. The broader evaluation dimensions outlined in Table \ref{tab_1} require deeper refinement, particularly for tele-traversability's downstream role, for instance, how it influences decision-making, control, human-robot coordination, and task performance.


\section*{Acknowledgments}
This work was supported by the National Natural Science Foundation of China, No. 52272411.

\bibliographystyle{unsrtnat}
\bibliography{reference}

\end{document}